# An Efficient Compression of Deep Neural Network Checkpoints Based on Prediction and Context Modeling

Yuriy L. Kim
*ITMO University*
Saint-Petersburg, Russia
ylkim@itmo.ru

Evgeny A. Belyaev
*ITMO University*
Saint-Petersburg, Russia
eabelyaev@itmo.ru

*Abstract*—This paper is dedicated to an efficient compression of weights and optimizer states (called checkpoints) obtained at different stages during a neural network training process. First, we propose a prediction-based compression approach, where values from the previously saved checkpoint are used for context modeling in arithmetic coding. Second, in order to enhance the compression performance, we also propose to apply pruning and quantization of the checkpoint values. Experimental results show that our approach achieves substantial bit size reduction, while enabling near-lossless training recovery from restored checkpoints, preserving the model's performance and making it suitable for storage-limited environments.

## I. Introduction

Efficient checkpoint compression has become increasingly important in recent years due to the growing number of parameters in deep learning models. A single checkpoint can take up tens of gigabytes, and there are many such checkpoints generated during model training. Therefore, General-purpose data compression methods can be applied to compress checkpoints. For instance, Prediction by Partial Matching [1] is a statistical-based method that employs a probability estimator and entropy coding, frequently using arithmetic coding for lossless data compression. Similarly, CMIX [2] uses 2,122 mixed models, including neural networks, to achieve high compression ratios across a wide range of domains. Several other methods have incorporated prediction models into the compression process. Tensorflow-compress [3], integrates LSTM networks with preprocessing steps from CMIX. Dzip [4] is a general-purpose compressor for sequential data that utilizes neural networks for prediction followed by arithmetic coding, featuring a hybrid architecture based on adaptive and semi-adaptive training. Additionally, TRACE [5] introduces a lossless compressor based on a single-layer transformer, utilizing byte-grouping and Shared-FFN schemes.

However, methods designed specifically for checkpoint compression tend to be more effective. LC-Checkpoint [6] proposes a lossy compression scheme for checkpoint construction, leveraging quantization, priority promotion, and Huffman coding. Delta-DNN [7] introduces a delta compression framework that exploits the similarity between neighboring versions of neural networks during training, using error-bounded lossy compression. QD-Compressor [8] builds on the ideas of Delta-DNN and LC-Checkpoint, employing a quantization-based delta compression approach with adaptive layer-wise quantization and an error feedback mechanism. Another approach, Inshrinkerator [9], utilizes a non-uniform quantization scheme, dynamic search for optimal quantization configurations, and a quantization-aware delta compression mechanism to enhance compression efficiency while preserving model accuracy. Nonetheless, existing methods primarily focus on compressing model weights alone, overlooking the need to retain additional information, such as optimizer states, required for correct training resumption in optimizers like Adam. Notably, the recent method ExCP [10] presents a framework for checkpoint compression, which addresses this limitation by compressing both the model weights and optimizer data, combining residual calculation, weight-momentum joint shrinking, and non-uniform quantization, what makes it currently the state-of-the-art checkpoint compression approach.

However, almost all existing compression methods store the quantized values without further processing them for better compression. ExCP uses the 7-zip archiver for this purpose, but this is not the most efficient approach. For further increasing the compression efficiency, in this paper, we introduce the assumption that there is a correlation between the quantized residual values of a reference checkpoint and the corresponding residuals of the current checkpoint. Based on this assumption, we propose using the neighboring values of the reference checkpoint as context for probability estimation via an LSTM [11] model, enabling more effective compression by capturing these correlations.

In order to use this information, in this paper we introduce a prediction-based compression approach. Specifically, during compression, we propose using the quantized values from the previous checkpoint as the input sequence to an LSTM model, which will predict the probability distribution of the values at the current checkpoint. This probability will then be used for encoding with an adaptive arithmetic coder [12]. Finally, our experimental results show that the proposed method provides a better compression ratio compared to other methods while maintaining the performance of the model after decompression, confirming that the assumption holds, and the proposed context modeling effectively improves compression efficiency.

The rest of the paper is organized as follows. Section II shortly describes the algorithm from ExCP, which is used as a basis for obtaining quantized values. Section III introduces the proposed compression method. Section IV provides comparisons with some compression methods. Section V concludes the presented results.

## II. Checkpoint Pruning and Quantization in ExCP

During the t-th training iteration, a neural network checkpoint $P_t$ consists of model weights $W_t$ and optimizer momentum parameters $O_t$:

$$Pt = \{Wt, Ot\}. \quad (1)$$

Given T checkpoints during training, the series of checkpoints is:

$$P = \{P1, P2, ..., PT\}. \quad (2)$$

Optimizer momentum parameters, particularly the first-order and second-order moments ($v_t$ and $m_t$), dominate storage. Thus, both weights and momentum states also considered for compression. Weights are stored as differences between adjacent checkpoints to leverage sparsity, while momentum states remain unchanged:

$$\Delta P_t = \{W_t - W_{t-1}, O_t\}. \quad (3)$$

To optimize compression, a joint pruning approach is utilized for both weights and momentum states. Residual weights are pruned using a threshold calculated based on the second-order moment $m_t$, which reflects the statistical average of weight changes:

$$r_w = \alpha / sqrt(m_t) \cdot median(W), M_w(i) = 1, if\ w_t(i) > r_w, \quad (4)$$

where W, $m_t$, $M_w$ represent the weights, the second-order moment, and the binary mask for weight pruning, respectively, and α is a hyperparameter. First-order momentum $v_t$ is used as an indicator for pruning momentum states. Momentum parameters corresponding to pruned weights are also discarded as:

$$r_o = \beta \cdot mean(v_t), M_o(i) = 1, if\ v_t(i) > r_o\ and\ M_w(i)=1, \quad (5)$$

where β is a hyperparameter, and $M_o$ is the binary mask for momentum prunin.

Subsequently, non-uniform quantization is employed to compress checkpoints further. The pruned values are set to zero, and the remaining parameters are clustered using the k-means algorithm to $2^n$ -1 cluster centers. These clusters are stored as indices and centers. Additionally, multiple lower-precision numbers, such as int4 or int2, are combined into a single higher-precision number (e.g., int8) during the saving process to optimize storage.

### III. Proposed Checkpoint Compression Method

In this paper, we introduce a prediction-based compression approach applied to the quantized values obtained using the ExCP method described in Section II. Fig. 1 shows an example of residual values from two adjacent checkpoints. It can be observed that the images exhibit structure and spatial correlation. Based on this, we propose using the residuals of the previous layer as context for prediction. Specifically, the algorithm works as follows: the quantized weight values are passed into the compressor, where the context for each weight is determined. The context includes the weight from the previous checkpoint and its surrounding weights, Fig. 2 shows the context formation diagram. This context is fed into an LSTM model, which computes a probability vector for the current weight based on the provided context. The size of the probability vector corresponds to the alphabet size, which is equal to $2^n$, where $n$ is the number of bits used for quantization. The actual value of the weight is then encoded using adaptive arithmetic coding. The entire processing occurs in batches. After each weight in batch is processed, the LSTM model is updated to reflect the new context. This process is repeated for all weights.

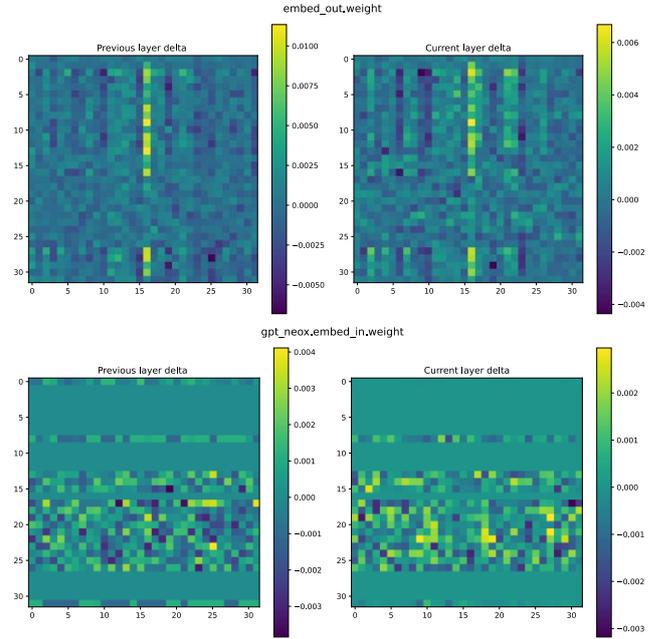

Fig. 1. An example of weight residuals correlation. On the left is the previous checkpoint, and on the right is the current one.

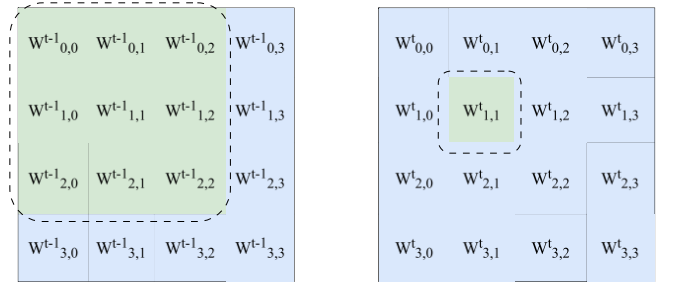

Fig. 2. An example of context for weight. On the left is the checkpoint t-1, and on the right is the checkpoint t.

During decoding, adaptive arithmetic decoding is used to decode the residuals, followed by predicting the weight using the LSTM model. The model is updated iteratively in the same manner as in encoding. Since the decoder operates symmetrically to the encoder, there is no need to transmit model parameters between encoding and decoding phases.

### IV. Perfomance Evaluation

For our experiments we replicated the setup from ExCP, utilizing the ViT-L32 [13] and Pythia-410M [14] models. The ViT-L32 model was trained and evaluated on the ImageNet-1K dataset, while the Pythia-410M model was trained on a subset of the standard Pile [15] dataset. During training, checkpoints were saved after each epoch for ViT-L32 and every 1000 iterations for the Pythia-410M, with compression applied simultaneously. The training process was interrupted periodically, and then resumed from compressed checkpoints. The period of breaking is set to 5000 iterations for Pythia-410M. Since the proposed compression method is lossless, the quality metrics obtained after compression are identical to those reported in ExCP.

The proposed method is implemented in PyTorch, ensuring compatibility with GPU acceleration for efficient execution. The LSTM model is trained with the following hyperparameters: batch size = 256, sequence length = 9, hidden units = 512, num layers = 2, embedding dimension = 512, learning rate = 0.001.The Adam optimizer [16] is used with hyperparameters β1 = 0, β2 = 0.9999, and ε = 1e-5, making it equivalent to RMSProp with a bias correction.

To evaluate the efficiency of the proposed compression method, we analyze the size of compressed checkpoints as the number of training iterations increases. The experiments are conducted on the Pythia-410M model, with checkpoints saved every 1000 iterations. The results are presented for three setups: ExCP, the proposed method, and the proposed method where the context is replaced by zero, similar to context-free probability estimation in arithmetic coder.

The graph in Fig. 3 shows the compressed checkpoint size (y-axis) as a function of training iterations (x-axis). The proposed method achieves a higher compression ratio, with compression increasing by an average of 14%. The compression ratio of the proposed method is around 90. After resuming training from the restored checkpoint, an increase in the size of the next checkpoint is observed, followed by a decrease in size during subsequent training. This is due to a reduction in the number of changes between checkpoints, which results in increased correlation.

In the next experiment, we test the effect of varying the step size for residuals calculation on the compression efficiency of the proposed method. The goal is to investigate whether omitting some checkpoints during training can lead to greater memory savings without significantly affecting the compression performance. The experiments were conducted on the ViT-L32 model, and we compared the compression performance when residuals were calculated from checkpoints spaced at different intervals:

$$\Delta P_t = \{W_t - W_{t-s}, O_t\}, \qquad (6)$$

where s is a step size.

The results shown in Fig. 4 highlight that, as in the previous experiment, compression increases with the number of iterations. Initially, the proposed method does not achieve better compression, but as training progresses, compression increases up to 31%. The compression ratio of the proposed method is around 50. Additionally, experiments were conducted with s = 2, so the reference checkpoint was not the previous one, but the one before that. This checkpoint merging approach could help save even more memory resources.

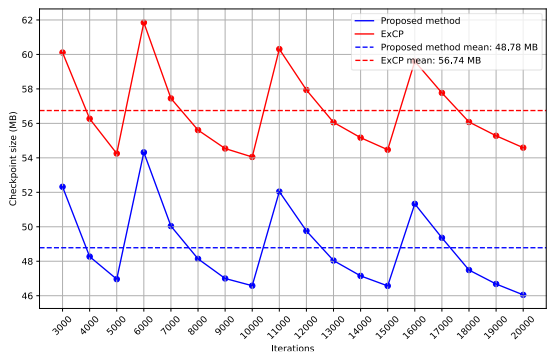

Fig. 3. Compressed checkpoint size as a function of training iterations for the proposed method and ExCP.

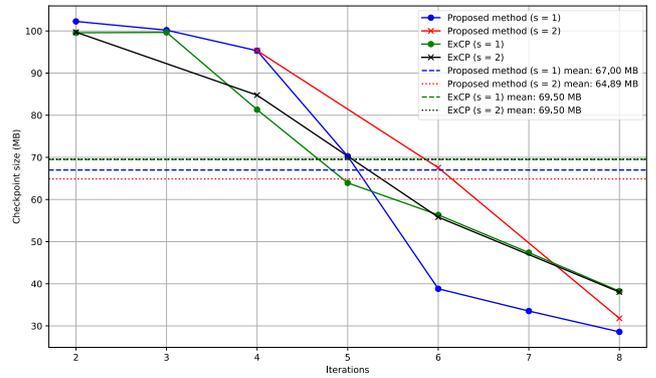

Fig. 4. Compressed checkpoint size as a function of training iterations for different step size s.

## V. CONCLUSION

In this paper we introduced prediction-based compression approach for model checkpoints. We have shown that the proposed method provides better compression ratio than existing checkpoint compressions frameworks, while maintaining model performance. Therefore, it could be highly beneficial for training models with a large number of parameters, helping to reduce resource usage.